\documentclass[pdflatex,sn-mathphys-num]{sn-jnl}% Vancouver Reference Style
\usepackage{anyfontsize}
\usepackage{amsthm}%
\usepackage{graphicx}%
\usepackage{multirow}%
\usepackage{amsmath,amssymb,amsfonts}%
\usepackage{changepage} % in preamble
\usepackage{mathrsfs}%
\usepackage[title]{appendix}%
\usepackage{textcomp}%
\usepackage{manyfoot}%
\usepackage{booktabs}%
\usepackage{algorithm}%
\usepackage{algorithmicx}%
\usepackage{algpseudocode}%
\usepackage{listings}%
\usepackage{todonotes}
\usepackage{subcaption}
\usepackage{tcolorbox}
\usepackage{setspace}
\usepackage{enumitem}
\usepackage{longtable}
\usepackage{wrapfig}
\usepackage{float}
\usepackage[dvipsnames]{xcolor}
\usepackage{pifont}
\newcommand\crule[3][black]{\textcolor{#1}{\rule{#2}{#3}}}
\newcommand{\highlight}[1]{\textcolor{black}{#1}} % changes to red
\newcommand{\sjh}[1]{\textcolor{black}{#1}} % changes to red
\newcommand{\highlightc}[1]{\textcolor{black}{#1}} % changes to red

\definecolor{color1}{HTML}{ff7f0e}
\definecolor{color2}{HTML}{d62728}
\definecolor{color3}{HTML}{2ca02c}
\definecolor{color4}{HTML}{1f77b4}
\definecolor{color5}{HTML}{f9ebea}

\newtcolorbox{descriptiontable}{
    colframe=white,    
    coltext=black,     
    arc=0pt,           
    width=1.3\textwidth, 
    boxrule=0.1pt,     
    boxsep=1pt,        
    fonttitle=\bfseries, % Bold title
    sharp corners=south, % Optional: for flat bottom
}
\newtcolorbox{descriptiontable2}{
    colframe=white,    
    coltext=black,     
    arc=0pt,           
    width=\textwidth, 
    boxrule=0.1pt,     
    boxsep=1pt,        
    fonttitle=\bfseries, % Bold title
    sharp corners=south, % Optional: for flat bottom
}

\begin{document}

\title[Explainable AI for Infection Prevention and Control: Modeling CPE Acquisition and Patient Outcomes in an Irish Hospital]{Explainable AI for Infection Prevention and Control: Modeling CPE Acquisition and Patient Outcomes in an Irish Hospital with Transformers}


\author*[1,2]{\fnm{Minh-Khoi} \sur{Pham}}\email{minhkhoi.pham@adaptcentre.ie}

\author[1,2]{\fnm{Tai} \sur{Tan Mai}}
\author[1,2]{\fnm{Martin} \sur{Crane}}
\author[2,3]{\fnm{Rob} \sur{Brennan}}
\author[4]{\fnm{Marie} \spfx{E.} \sur{Ward}}
\author[4]{\fnm{Una} \sur{Geary}}
\author[4]{\fnm{Declan} \sur{Byrne}}
\author[4]{\fnm{Brian} \sur{O’Connell}}
\author[4,6]{\fnm{Colm} \sur{Bergin}}
\author[5]{\fnm{Donncha} \sur{Creagh}}
\author[6]{\fnm{Nick} \sur{McDonald}}
\author[1,2]{\fnm{Marija} \sur{Bezbradica}}

\affil[1]{\orgdiv{School of Computing}, \orgname{Dublin City University}, \orgaddress{\state{Dublin}, \country{Ireland}}}
\affil[2]{\orgname{ADAPT Centre}, \state{Dublin}, \country{Ireland}}
\affil[3]{\orgname{University College Dublin}, \state{Dublin}, \country{Ireland}}
\affil[4]{\orgname{St James's Hospital}, \state{Dublin}, \country{Ireland}}
\affil[5]{\orgname{Health Service Executive (HSE)}, \state{Dublin}, \country{Ireland}}
\affil[6]{\orgname{Trinity College Dublin}, \state{Dublin}, \country{Ireland}}


\abstract{
\highlightc{\textbf{Background:} Carbapenemase-Producing Enterobacteriace (CPE) poses a critical concern for infection prevention and control in hospitals. However, predictive modeling of previously highlighted CPE-associated risks such as readmission, mortality, and extended length of stay (LOS) remains underexplored, particularly with modern deep learning approaches. This study introduces an eXplainable AI (XAI) modeling framework to investigate CPE impact on patient outcomes  from Electronic Medical Records (EMR) data of an Irish hospital.}

\highlight{\textbf{Methods:} We analyzed an inpatient dataset from an Irish acute hospital (2018-2022), incorporating diagnostic codes, ward transitions, patient demographics, infection-related variables and contact network features. Several Transformer-based architectures (e.g., TabTransformer, TabNet) were benchmarked alongside traditional machine learning models. Clinical outcomes were predicted, and XAI techniques were applied to interpret model decisions.}

\highlight{\textbf{Results:} Our framework successfully demonstrated the utility of Transformer-based models, with TabTransformer consistently outperforming baselines across multiple clinical prediction tasks, especially for CPE acquisition (Area Under Receiver Operating Characteristic and sensitivity). We found infection-related features, including historical hospital exposure, admission context, and network centrality measures, to be highly influential in predicting patient outcomes and CPE acquisition risk. Explainability analyses revealed that features like \texttt{"Area of Residence"}, \texttt{"Admission Ward"} and prior admissions are key risk factors. Network variables like \texttt{"Ward PageRank"} also ranked highly, reflecting the potential value of structural exposure information.}
% Isolation and screening indicators provided variable but insightful signals, which XAI methods effectively surfaced for individualized review.}

\highlight{\textbf{Conclusion:} This study presents a robust and explainable AI framework for analyzing complex EMR data to identify key risk factors and predict CPE-related outcomes. Our findings underscore the superior performance of the Transformer models and highlight the importance of diverse clinical and network features. The transparent interpretability offered by our XAI approach provides actionable insights for infection prevention and control, paving the way for more targeted interventions and ultimately enhancing patient safety within acute healthcare settings.}
}




\maketitle

\section{Background}

Infection Prevention and Control (IPC) is a cornerstone of modern healthcare, aiming to reduce the spread of healthcare-associated infections (HCAIs) and mitigate their significant impact on patients, healthcare professionals, and hospital systems. Among the most urgent threats in this domain are Carbapenemase-Producing Enterobacterales (CPE)—a class of multidrug-resistant organisms that diminish treatment efficacy in clinical environments \cite{codjoe2017carbapenem, haque2018health}. The World Health Organization formally recognizes HCAIs as a critical patient safety concern through its International Classification for Patient Safety (ICPS) \cite{safety2010conceptual}. In Ireland, CPE has emerged as a national health priority, with case numbers rising steadily over the past decade \cite{vellinga2021initial, o2022microbial, humphreys2022reflections, fahy2020carbapenemase}. This growing burden underscores the need for smarter surveillance, screening strategies, and data-driven models to support timely IPC interventions \cite{kardas2022cost, fahy2020carbapenemase}.

This work is part of the ARK-Virus project, an interdisciplinary initiative focused on advancing IPC practices across Irish acute hospitals \cite{ward2024systems}. By combining expertise in microbiology, human ergonomics, informatics, and data science, the project seeks to develop risk-aware, evidence-based frameworks to support healthcare decision-making \cite{mcdonald2021evaluation, crotti2020ark}. Within this framework, our work focuses on using AI technologies to strengthen IPC for CPE infections, leveraging longitudinal data from Electronic Medical Records (EMRs) and applying machine learning (ML) to investigate how transmission risk, screening strategies, and patient characteristics intersect with clinical outcomes \cite{pham2024forecasting}.

\sjh{CPE infections are known to associate with poor patient outcomes, including prolonged hospital stays, higher in-hospital mortality, and increased risk of readmission \cite{corson2020hospital, emerson2012healthcare}. In this study, we aim to quantify these impacts more systematically by modeling three key clinical endpoints: 30-day readmission, in-hospital mortality, and length of stay (LOS). By leveraging AI, we aim to gain actionable insights into the correlation and causation between CPE infection and the patient outcomes, informing both clinical risk stratification and hospital resource planning.}

\sjh{Recent advances in AI have significantly contributed to IPC, especially in the surveillance and prediction of HCAIs, antimicrobial resistance (AMR), and timely intervention planning \cite{fitzpatrick2020using, yang2024implications, abu2025artificial}. Numerous models have been developed to forecast colonization risks of multidrug-resistant organisms at hospital admission by leveraging features such as prior antibiotic use, comorbidities, invasive procedures, and long-term care exposure \cite{caglayan2022data, wang2023deep}. Contact network-based features—derived from ward co-location and patient trajectories have also been explored to model in-hospital transmission risk \cite{myall_prediction_2022, gouareb2023detection}, while large-scale EMR-based models like ARGai 1.0 have demonstrated the utility of Transformer-based architectures in tasks ranging from gene expression classification to pathogen detection \cite{nayak2025argai}. However, specific attention to CPE remains limited, especially within the Irish healthcare context. Prior studies on CPE typically focus on individual risk factors \cite{segagni2020infection, goodman2019predicting, freire2022prediction, liang2024prediction, liang2022early}, but rarely integrate contact data or interpretability frameworks. We summarize the known CPE risk factors reported in the literature in Table~\ref{tab:cpe_risk_factors}.}

\begin{table}[htbp]
\centering
\caption{Common Risk Factors for CPE Acquisition Identified in Literature}
\label{tab:cpe_risk_factors}
\begin{tabular}{p{0.8\linewidth} r}
\toprule
\textbf{Risk Factor} & \textbf{References} \\
\midrule
Prior antimicrobial use (carbapenems, cephalosporins, fluoroquinolones) & \cite{segagni2020infection, european2011risk} \\
Recent hospitalization (within past year) & \cite{segagni2020infection, goodman2019predicting} \\
Prolonged hospital stay ($>$20 days) & \cite{european2011risk, liang2024prediction} \\
Stay in high-risk wards (e.g., ICU, transplant units) & \cite{segagni2020infection, caglayan2022data} \\
Use of invasive devices (e.g., catheters, ventilation) & \cite{wang2023deep, liang2024prediction} \\
Surgical interventions & \cite{segagni2020infection} \\
Comorbidities (e.g., diabetes, CKD, malignancy) & \cite{freire2022prediction, wang2023deep} \\
Cross-border healthcare exposure or patient transfers & \cite{european2011risk, goodman2019predicting} \\
Residence in long-term care facilities & \cite{caglayan2022data, gouareb2023detection} \\
Severity of illness or immunosuppression & \cite{freire2022prediction, wang2023deep} \\
\bottomrule
\end{tabular}
\end{table}

Transformer-based models such as Med-BERT \cite{rasmy2021med}, BEHRT \cite{li_behrt_2020}, \highlight{Hierarchical BEHRT} \cite{li_hi-behrt_2023}, CORE-BEHRT \cite{odgaard2024core}, and EHR-Mamba \cite{fallahpour2024ehrmamba} have shown strong performance for EMR-based patient outcome prediction. These models leverage specialized embeddings for age, visit gaps, and diagnoses, often pretrained on large-scale EMR corpora and fine-tuned for downstream clinical tasks. Tabular deep models like \highlight{Convolutional Networks \cite{bednarski2022temporal}}, TabNet \cite{arik2021tabnet}, TabTransformer \cite{huang2020tabtransformer}, ResNet \cite{gorishniy2021revisiting}, and TabPFN \cite{hollmann_accurate_2025} have similarly demonstrated effectiveness across general tabular datasets, but not quite well studied in healthcare domain. In parallel, explainability has become a critical component in model adoption. While tools like SHAP and LIME are commonly used for global interpretation \cite{loh2022application, di2023explainable, chaddad2023survey}, \highlight{techniques like DeepXplainer \cite{wani2024deepxplainer} and recommendations by Wani et al. \cite{wani2024explainable} highlight the need for interpretable models in clinical environments}. Recently, gradient-based approaches such as Integrated Gradients have been used to gather finer insight into deep model behavior \cite{sundararajan2017axiomatic}, even for medical models \cite{fallahpour2024ehrmamba}. Our study builds upon this work by benchmarking recent tabular and EMR-specific models for CPE outcome prediction, embedding contact network features, and applying integrated explainable AI (XAI) tools to support actionable insights.

\highlight{To model these phenomena, we use a real-world anonymized EMR dataset from an Irish acute hospital spanning four years of inpatient admissions. We employ a mix of machine learning models—ranging from traditional gradient-boosted trees to explainable deep learning approaches such as ResNet, TabNet and Transformer-based architectures—to capture complex temporal and contextual patterns that influence CPE risk and outcomes. More importantly, we introduce \textbf{contact network-derived features} based on patient room overlaps and exposure histories to model potential transmission pathways within the hospital. Given the critical nature of care in such a setting and known challenges of deep model transparency, we emphasize explainability throughout, applying multiple XAI techniques to understand feature contributions and to build trust in model-driven IPC strategies.}

\vspace{0.5em}
\noindent
\textbf{Objective.} This work presents a domain-centered, explainable AI modeling framework for investigating CPE impact on clinical outcomes in an Irish hospital. Rather than proposing novel architectures, we aim to systematically apply and evaluate existing state-of-the-art Transformer-based models in a real-world healthcare setting, with a focus on IPC-relevant insights.

\vspace{0.5em}
\noindent
\textbf{Research Questions.} This study is guided by three central research questions:
\begin{itemize}
    \item \textbf{(RQ1)} How do infection-related features impact on future patient outcomes such as 30-day readmission, in-hospital mortality, and LOS?
    \item \textbf{(RQ2)} Which patient features are the most predictive of CPE infection risk?
    \item \textbf{(RQ3)} Can deep learning and XAI methods identify distinctive patterns among CPE-positive patients at the cohort level, and how can these tools support personalized analysis at the individual level?
\end{itemize}

\vspace{0.5em}
\noindent
\textbf{Contributions.} The main contributions of this study are:
\begin{itemize}
    \item We benchmark a suite of traditional, deep, and Transformer-based models on a rare and newly gathered CPE dataset, offering comparative performance insights.
    \item We incorporate contact network-derived features to model in-hospital exposure, supporting nuanced analysis of CPE acquisition dynamics.
    \item We utilize Integrated Gradients  to study both global and individual-level patterns of infection and risk. With this, we generate clinically actionable findings on the potential operational effectiveness of screening and isolation strategies, directly supporting IPC optimization.
\end{itemize}

% \input{Backups/related_work}

%%%%%%%%%%%%%%%%%%%%%%%%%%%%%%%%%%%%%%%%%%

\section{Dataset}
\label{sec:datasets}

\subsection{Dataset description}
\label{sec:datadesc}

We utilize anonymized inpatient records from a national acute hospital in Ireland, collected between January 2018 and February 2022 as part of the ARK-Virus project \cite{ward2024systems}. \highlightc{The dataset is derived from multiple sources including the national Hospital Inpatient Enquiry (HIPE) system and Patient Admission System from the hospital discharge database of an Irish acute hospital~\cite{o2005using, hipe2020v14}, and includes nearly one million bed-day records.}

\begin{figure}[ht]
    \centering
    \includegraphics[trim={0.4cm 0.2cm 2cm 2.3cm},clip, width=\linewidth]{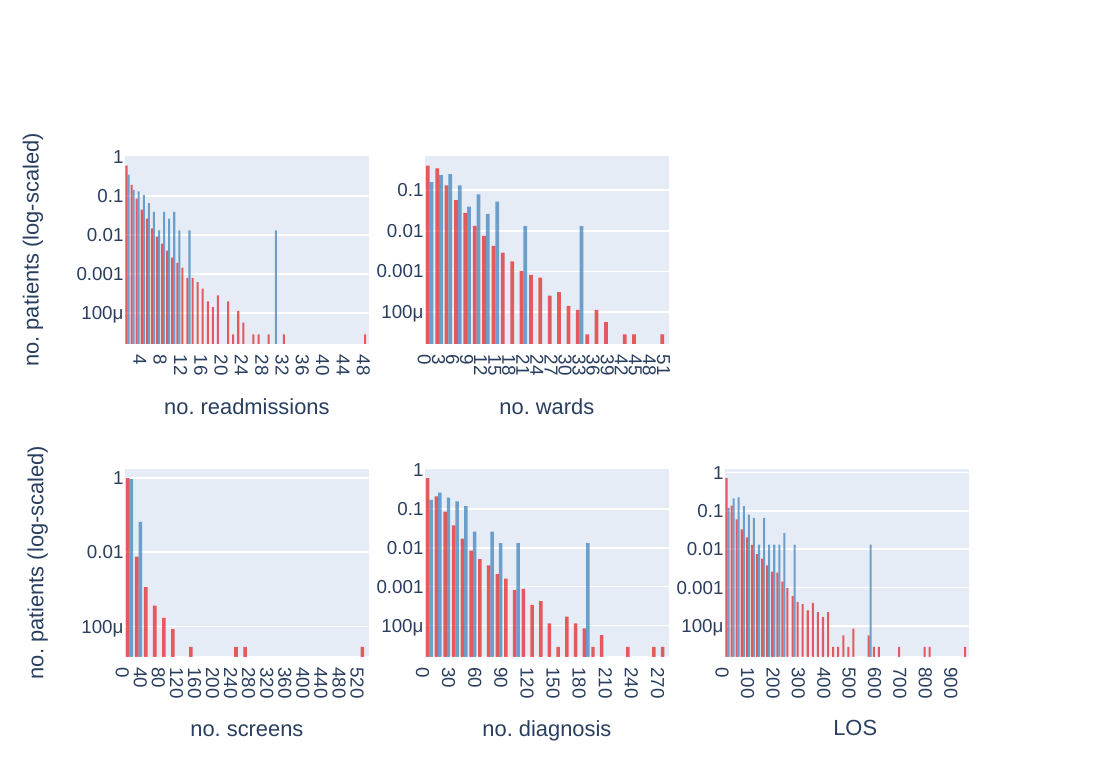}
    \caption{Distribution of the two studied patient cohorts based on CPE screening status\highlightc{, whether it is positive (CPE+) or negative (CPE-)}, \crule[color2]{0.2cm}{0.2cm}: CPE-, \crule[color4]{0.2cm}{0.2cm}: CPE+}.
    \label{fig:distribution}
\end{figure}

Each entry contains structured metadata on patient demographics, admission and discharge details, ward transfers, and International Classification of Diseases (ICD-10) codes for diagnoses and procedures \cite{world1992icd}. \highlight{Additionally, the dataset contains results from CPE screening tests conducted during hospitalization. Most CPE-positive cases in our dataset are identified through these routine screenings, which primarily detect colonization (organism presence without symptoms) rather than active infection. As shown in Figure \ref{fig:distribution}, CPE-positive patient cohort seems to exhibit slightly higher numbers of readmissions, ward transitions, diagnoses, and longer lengths of stay compared to CPE-negative patients. Full summary statistics of the data and a comprehensive overview of the feature types included is presented in Appendix.}

\subsection{Cohort Construction and Preprocessing}
\label{section:cohort_construction}
We constructed our study cohort by applying a multi-stage filtering and preprocessing pipeline:

\sjh{\begin{enumerate}
    \item Anonymized data was shared on all acute inpatient admissions between January 1, 2018 and February 28, 2022.
    \item Data on adult patients (age $\geq$ 18) with hospital stays of at least 48 hours and at least one prior admission.
    % , to ensure sufficient observation time for potential CPE colonization or acquisition was retained.
    \item CPE status was assigned based on the presence of a positive screening test during the admission. In-hospital mortality was identified using the discharge status field.
    \item To simulate prospective evaluation, we performed a chronological data split, allocating 90\% of episodes to the training set and 10\% to the test set.
\end{enumerate}}

\noindent
\highlight{To reduce diagnostic code sparsity while retaining clinical meaning, we applied code generalization: diagnosis codes were truncated to the first three digits of the ICD-10 hierarchy, and procedure codes to the first five digits. This approach aligns with prior research on medical code abstraction \cite{deschepper2019hospitalwide, wang2018predictingclinical, choi2016doctor}. Missing entries were encoded as \texttt{"None"}, and duplicate rows or empty columns were removed to ensure data integrity.}

\subsection{Feature Engineering}
\label{sec:feature_engineering}

\subsubsection{Past Episodes Features}
\label{sec:methods:history}

Patient history plays a critical role in personalized risk modeling. Studies show that incorporating visit-level sequences of clinical events enhances predictive performance \cite{allam2021analyzing, choi2017using, pham2017predicting}. While our dataset lacks fine-grained timestamps, we extract cumulative historical features (e.g, such as the total number of diagnoses, comorbidities, ward transfers, and prior admissions) to summarize past medical activities. This approach is expected to maintain the predictive utility of longitudinal data while being compatible with both static tabular models and Transformer-based architectures. 
 
\subsubsection{Ward transitions Features}

\sjh{During a patient’s hospital stay, following diagnosis, they may be transferred between different medical wards to receive specialist care or for bed management reasons. According to the research we gathered in Table \ref{tab:cpe_risk_factors}, ward information can contain useful information that are indicative of CPE infection. To capture this, we compute the frequency of admissions to both admission and discharge wards, which the model encodes as numerical and categorical features, respectively. These features are presented in the Appendix as ``\texttt{No. Acute Department visits}", ``\texttt{No. Emergency Department visits}", ``\texttt{Admission Ward}", and ``\texttt{Discharge Ward"}}

\subsubsection{CPE-related Features}

\sjh{CPE screening data plays a central role in our analysis. We extracted variables indicating whether a patient underwent CPE testing during an admission, as well as the corresponding results (positive/negative). In some cases, diagnosis codes also contained CPE-related entries. Screening is conducted in accordance with Irish Health Service Executive \cite{hse20xxCPE}, typically for surveillance or diagnostic purposes.} \sjh{These screening variables serve as outcome labels in our CPE acquisition modeling (in Section \ref{sec:results:cpe}), and as input features in our patient outcome prediction tasks (in Section \ref{sec:results:outcomes}). Their presence enables us to evaluate their influence on model predicting the clinical outcomes.}

\highlightc{Through out the paper, for the ease of writing, we use the term “CPE-positive” as a patient who has had at least one positive result on CPE screening, regardless of whether this reflects carriage or active infection. We note that infection status can change over time (e.g., clearance of infection or subsequent acquisition), while screening captures both colonization and infection events. This definition is adopted for consistency and interpretability within our modeling framework}

\subsubsection{Contact Network Features}
\label{sec:contact_features}

\sjh{To capture potential transmission dynamics within the hospital, we constructed daily patient contact networks at the ward level. Two patients were considered \emph{in contact} if they occupied the same ward on the same day. Any patient who shared a ward with a CPE-positive case was marked as \emph{exposed}.}
% We visualize the patient contact networks across the dataset timeline as the scatter plots in Figure \ref{fig:patient_contacts}. Each node represents a patient hospitalizing during that time, edges indicate whether two patients were \emph{in contact}.}

% \highlight{\begin{figure}[ht]
%     \centering
%     \includegraphics[trim={4cm 0 5cm 0},clip, width=\linewidth]{Figures/new/contact.pdf}
%     \caption{Visualization of CPE-positive patient contact networks over time. The histogram shows daily case counts, highlighting notable peaks in late 2018 and late 2021. Scatter plots illustrate sample daily contact networks. \crule[color1]{0.2cm}{0.2cm}: Exposed unscreened, \crule[color5]{0.2cm}{0.2cm}: Exposed screened, \crule[color2]{0.2cm}{0.2cm}: Infectious, \crule[color4]{0.2cm}{0.2cm}: Other unscreened.}
%     \label{fig:patient_contacts}
% \end{figure}}

\sjh{From these networks, we computed a set of graph-based features for each patient admission, including node degree, closeness centrality, and daily contact counts (formal definitions of these terms can be found in Appendix). These features were aggregated over the hospital stay. While prior studies have used finer spatial granularity (e.g., room- or building-level) \cite{myall_prediction_2022}, we limited our analysis to ward-level data due to availability constraints. We report the statistics of these features in the Appendix.}

\section{Methodology}
\label{sec:methods}

\subsection{Overview of the Framework}
\label{sec:methods:overall}

\begin{figure}[ht]
    \centering
    \includegraphics[trim={0cm 0cm 0cm 2cm},clip, width=\linewidth]{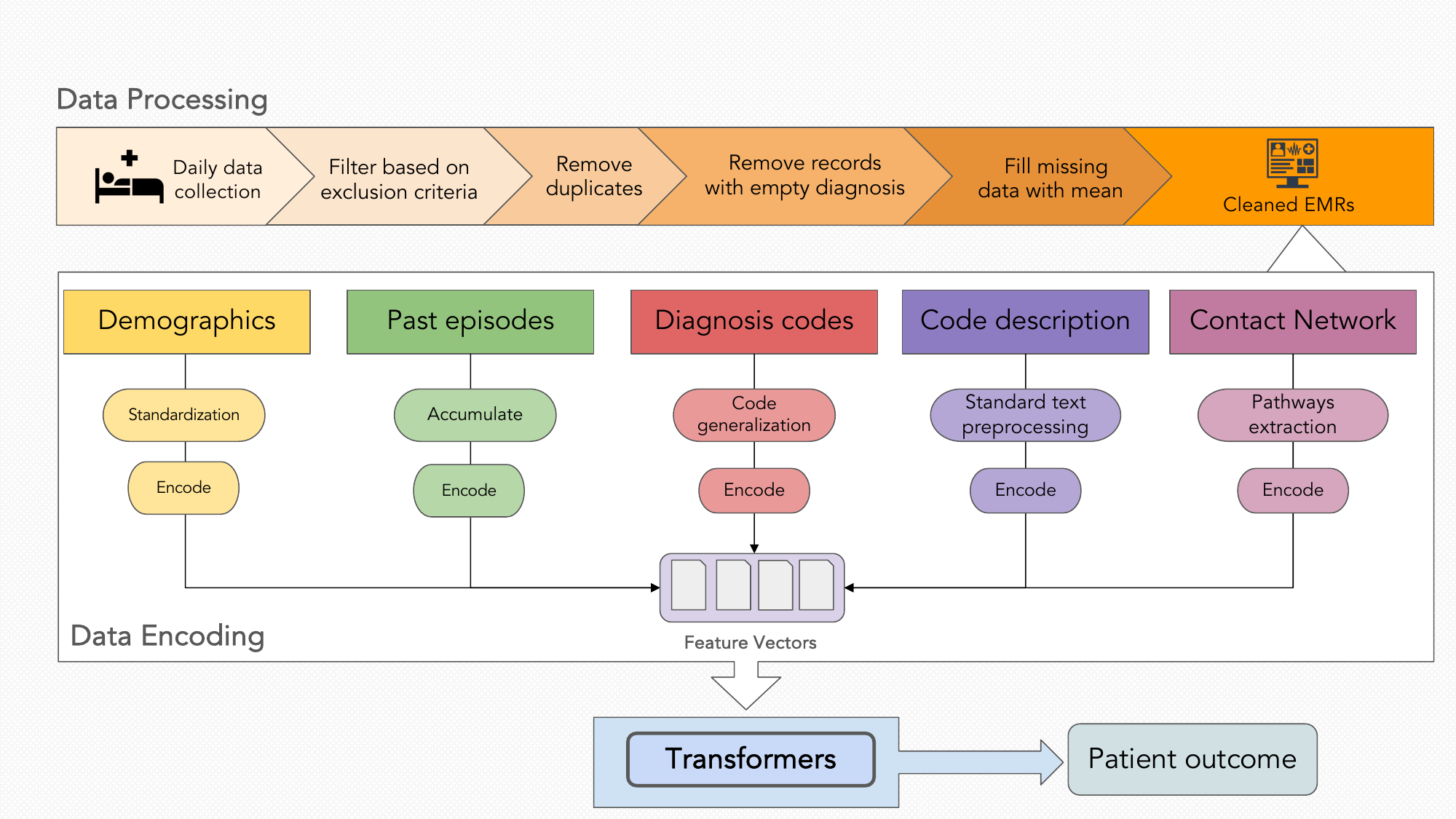}
    \caption{Overview of the proposed framework for processing and modeling structured EMR data. The pipeline integrates demographic data, ward history, diagnosis codes, and contact network features into Transformer-based predictive models.}
    \label{fig:framework}
\end{figure}

\highlight{Our approach integrates structured EMR data with contact network-derived features to forecast both clinical outcomes and infection risks. As shown in Figure~\ref{fig:framework}, we combine information from multiple sources including demographics, past episode summaries, ICD-10 codes (and their textual descriptions), and in-hospital patient contact networks. Each feature type is encoded separately according to its data modality.}

\highlight{For numerical features, we use multi-layer perceptron (MLP) layers to project them into the shared embedding space. Categorical variables are encoded via lookup tables to obtain dense vector embeddings. All projected representations are aligned into a common hidden dimension and concatenated before model fusion.}

\highlight{To preserve diagnostic history, we incorporate all diagnosis and procedure codes from both the index admission and prior hospitalizations. Prior work has demonstrated that leveraging longitudinal data in this way significantly improves predictive accuracy \cite{pham2024forecasting, segagni2020infection, goodman2019predicting}.}

\highlight{Clinical codes are among the most informative inputs for modeling patient trajectories and predicting outcomes \cite{moloney2006factors}. However, their high dimensionality and sparsity introduce significant challenges. To mitigate this, we perform code generalization (as described in Section \ref{section:cohort_construction} and utilize a publicly available pretrained embeddings \cite{wu2022comparison}. By doing this, the list of billing codes patients received during the admission are transformed into a list of feature vectors. These vectors are then, similarly, projected into the shared embedding space by another MLP layer. For the procedure codes, which lack pretrained resources, we train embedding layers from random initialization. }

 \highlight{To further address sparsity and enhance generalization, we enrich the clinical code representations with semantic embeddings derived from their textual descriptions. Specifically, we use a pretrained BioBERT model\footnote{we leveraged model from \texttt{huggingface.com/nlpie/tiny-biobert}} to generate 768-dimensional semantic vectors for each ICD code description. These vectors are then linearly projected to match the model's shared input dimension, allowing for integration with other feature types.}

The encoded features are processed through a model-specific backbone—such as ResNet, attention-based models, or Transformer variants—and subsequently projected into a unified embedding space. These combined representations are then passed through a MLP head for final prediction. \highlight{We benchmark several recent deep learning models tailored for structured or EMR data, each utilizing its own encoding and feature fusion strategy, as detailed in Section~\ref{sec:baseline_comparison}.}

\subsection{XAI techniques}
\label{sec:xai}

% \todo[inline]{
% - Provide more detailed explanations of XAI techniques implementation. 

% \\
% - Discuss how explanations were validated (consistency across folds, etc.)
% }

To deepen our understanding of model behavior and uncover patterns in the data, we incorporate a range of XAI techniques, selected to be compatible with the deep learning architectures used in this study. In particular, we focus on \textbf{Integrated Gradients (IG)} \cite{sundararajan2017axiomatic}, a theoretically grounded method for attributing prediction outcomes to input features. Despite its strong theoretical foundations, IG remains underutilized in clinical tabular modeling, as noted by Allgaier et al. \cite{allgaier2023does}.

Unlike SHAP, which relies on sampling an exponentially large space of possible feature combinations to approximate feature contributions \cite{shapley1953value}, IG computes gradients along a path from a baseline input to the actual input. This makes it both computationally efficient and well-suited for interpreting predictions from differentiable deep models. IG also enables flexible use cases, from estimating global feature importance across populations to individualized explanations at the patient level.

\highlight{To address our research questions—\textbf{identifying key predictors of patient outcomes}—we calculate feature attributions using IG across multiple benchmarked models (TabNet, TabTransformer and ResNet) and cross-validation folds. For each model, we compute the IG values relative to a baseline input, and then average the resulting scores across all samples. Feature importance rankings are derived by aggregating these scores across the results and ranked. We report and visualize the top contributing features per outcome, providing insights into which variables consistently influence model decisions in Section \ref{sec:results:xai}.}

% \todo[inline]{Provide in-depth analysis of why certain features appear more important than others}

% \todo[inline]{If possible, include feedback from clinicians on the explanations}

% \todo[inline]{Compare and contrast results from different XAI techniques (different baseline, different Feature Importance methods by Captum}

\section{Experimental Designs}

\subsection{Validation Method and Performance Metrics}

Many researchers have studied suitable evaluation metrics for classification tasks within the healthcare domain \cite{hicks2022evaluation}. They have emphasized the importance of not relying solely on a subset of metrics, as doing so could potentially yield misleading results when implementing models in clinical settings. To comprehensively assess model performance, we employ multiple evaluation metrics, including Sensitivity, Specificity, Area Under the Receiver Operating Characteristic Curve (AUROC) and Area Under the Precision Recall Curve (AUPRC). 

Specificity places a stronger emphasis on correctly classifying negative samples, while sensitivity aims to minimize the misclassification of positive instances, making it a critical metric in medical studies \cite{hicks2022evaluation}. \textit{AUROC} and \textit{AUPRC} has gained popularity in healthcare ML due to their favorable properties when dealing with imbalanced classes. They are used to distinguish between positive and negative classes across different threshold settings regardless of class distribution, making it an effective balance measurement. 

To assess the robustness of our models, we employ cross-validation techniques, particularly stratified 5-fold cross-validation. This allows consistent distribution of classes, ensuring stability in the benchmarking results. Following the guidance of \cite{hicks2022evaluation}, we avoid sharing data from the same patients across different folds to prevent introducing bias during the parameter tuning phase. We report the mean and standard deviation of model performance across these folds, and predictions made on the test set.

\subsection{Baseline comparison}
\label{sec:baseline_comparison}
We benchmark a broad set of Transformer-based architectures alongside traditional machine learning models tailored for tabular and EMR-based patient data. These models range from established tree-based ensembles to recently proposed neural architectures. The comparison results are summarized in Table~\ref{tab:benchmark_models}.

\vspace{0.5em}
\highlight{\noindent\textbf{XGBoost \cite{chen2016xgboost}, LightGBM \cite{ke2017lightgbm}, and CatBoost \cite{prokhorenkova2018catboost}.}
These gradient boosting frameworks serve as strong baselines for structured data. Patient records are represented using frequency-based vectors of categorical medical codes combined with aggregate statistics of numerical features.}

\vspace{0.5em}
\highlight{\noindent\textbf{ResNet.}
We implement a ResNet-based model following the tabular adaptation proposed in \cite{dai2022revisiting}, which demonstrates that residual connections in MLP architectures can achieve competitive performance. In their benchmarks, this approach sometimes outperforms more complex attention-based models.}

\vspace{0.5em}
\highlight{\noindent\textbf{TabNet \cite{arik2021tabnet}.} 
TabNet uses sequential attention to select salient features at each decision step, enabling both interpretability and efficient learning. Its encoder comprises feature transformers, attentive transformers, and a feature-masking mechanism. The decoder reconstructs inputs via a sequence of feature transformer blocks. Sparse attention via \textit{sparsemax} encourages the model to focus on a subset of input dimensions, providing interpretable insights into feature relevance.}

\vspace{0.5em}
\highlight{\noindent\textbf{TabTransformer \cite{huang2020tabtransformer}.} TabTransformer leverages self-attention to generate contextual embeddings for categorical features before combining them with continuous inputs in a downstream MLP. Its column-wise Transformer encoder is particularly effective in modeling high-cardinality categorical data, which is common in EMRs.}

\vspace{0.5em}
\highlight{\noindent\textbf{TabPFN \cite{hollmann_accurate_2025}.} 
TabPFN is a few-shot tabular classifier that emulates Bayesian inference using a Transformer trained entirely on synthetic data. It directly outputs calibrated class probabilities without requiring fine-tuning. Though originally designed for small-scale datasets, we include it as a fast and lightweight baseline to assess how well strong priors generalize to real-world healthcare data, despite its limit of 500 features and 10,000 training samples.}

\vspace{0.5em}
\highlight{\noindent\textbf{CoreBEHRT \cite{odgaard2024core}.}  
CoreBEHRT builds on the BEHRT architecture by modeling visit-level sequences using age embeddings, time gaps, and discretized medical values. While powerful for temporal EMR data, our application context is limited by coarse-grained (bed-day level) timestamps and the absence of lab or medication data. Consequently, CoreBEHRT only utilizes clinical code sequences, which may restrict its effectiveness compared to other models with richer feature sets.}

\subsection{Handling Class Imbalance}
\highlight{Patient clinical outcomes, such as 30-day readmission or mortality or CPE-related outcomes, are often rare, leading to highly imbalanced datasets. To mitigate this issue, we employed a range of training strategies:}

\highlight{\begin{itemize}
\item \textbf{Focal Loss} \cite{lin2017focal} is used as the primary classification loss, which down-weights easy negatives and focuses learning on hard, minority-class examples.
\item \textbf{Random Oversampling} is applied during training, as shown to be effective in our previous work \cite{pham2024forecasting}, ensuring minority-class samples are adequately represented.
\item \textbf{Balanced Batch Sampling} is used to maintain class parity within mini-batches, ensuring that gradients are not dominated by the majority class, especially helpful for deep neural networks.
\end{itemize}}

\vspace{0.5em}
\noindent
\highlight{These methods help prevent the model from becoming biased toward negative (non-readmission or non-mortality) outcomes and improve sensitivity to rare events. Evaluation across folds confirms that these strategies contribute to more stable training and improved detection of high-risk patients.}

% %%%%%%%%%%%%%%%%%%%%%%%%%%%%%%%%%%%%%%%%%%
\section{Results}
\label{sec:experiment} 

\subsection{Quantitative results}
\subsubsection{Patient Outcome forecast}
\label{sec:results:outcomes}
\highlight{We evaluate a broad set of baseline and Transformer-based models on three clinically relevant outcome prediction tasks: 30-day readmission, in-hospital mortality, and length of stay (LOS). Results are summarized in Table~\ref{tab:benchmark_models}. All models were trained using cross-validation and evaluated on a held-out test set, ensuring consistent comparisons across architectures.}

\begin{table*}[ht]
    \centering
    \caption{Benchmark performance of baseline and Transformer-based models on three clinical outcome prediction tasks using AUROC $/$ AUPRC. The results are trained using cross-validation and benchmarked on our data testset.}
    \label{tab:benchmark_models}
    \resizebox{\textwidth}{!}{%
    \begin{tabular}{|l|cc|cc|c|}
        \hline
        \multirow{2}{*}{\textbf{Model}} & 
        \multicolumn{2}{|c|}{\textbf{30-day Readmission}} &
        \multicolumn{2}{|c|}{\textbf{In-Hospital Mortality}} &
        \textbf{Length of Stay} \\
        &  \textbf{AUROC} $\uparrow$ & \textbf{AUPRC} $\uparrow$ & \textbf{AUROC} $\uparrow$ & \textbf{AUPRC} $\uparrow$ & \textbf{RMSE} $\downarrow$ \\
        \hline
        XGBoost                     & 0.824 (0.010) & 0.432 (0.020) & 0.702 (0.032) & \underline{0.024} (0.005) & 20.328 (2.306) \\
        CatBoost                     & 0.827 (0.005) & 0.446 (0.015) & 0.714 (0.015) & \underline{0.024} (0.028) & 16.066 (0.170)  \\
        LightGBM                  &  0.814 (0.055) & 0.438 (0.065) &  0.713 (0.028) & \underline{0.024} (0.001) & 16.703 (0.286) \\
        TabNet \cite{arik2021tabnet}   & 0.827 (0.005) & 0.440 (0.009) & \textbf{0.732} (0.006) & \underline{0.024} (0.001) & 54.766 (26.21) \\
        ResNet \cite{gorishniy2021revisiting}  & \underline{0.830} (0.001) &  \underline{0.457} (0.006) & 0.697 (0.006) & 0.023 (0.001) & 16.161 (0.086) \\
        TabPFN \cite{hollmann_accurate_2025}  & 0.828 (0.002) & 0.455 (0.004) & 0.700 (0.010) & \underline{0.024} (0.003) & \underline{15.713} (0.018) \\
        TabTransformer \cite{huang2020tabtransformer}  & \textbf{0.834} (0.004)  &  \textbf{0.461} (0.007) & \underline{0.728} (0.022) & \underline{0.024} (0.003) & \textbf{15.463} (0.109) \\
        CoreBEHRT \cite{odgaard2024core}       & 0.678 (0.001) & 0.309 (0.003) & 0.619 (0.030) & 0.017 (0.001) & 46.82 (0.509) \\
        \hline
    \end{tabular}
    }
\end{table*}

\highlight{As can be seen from the table, Transformer-based models (apart from CoreBEHRT) consistently outperform traditional gradient boosting methods (XGBoost, CatBoost, LightGBM) across nearly all tasks. In particular, \textbf{TabTransformer} achieves the best overall performance—achieving the highest AUROC and AUPRC for 30-day readmission and the lowest RMSE for length-of-stay prediction. These results highlight its ability to effectively capture complex tabular dependencies in EMRs.}

\highlight{\textbf{TabNet} also demonstrates competitive performance, particularly on the in-hospital mortality task, where its attention-driven feature selection appears well-suited for distinguishing high-risk patients. \textbf{ResNet} and \textbf{TabPFN} follow closely behind, with \textbf{TabPFN} showing impressive results on LOS prediction despite its data constraints (limited to 10,000 samples and 500 features), indicating strong potential in resource-constrained settings.}

\highlight{In contrast, \textbf{CoreBEHRT} underperforms across all tasks, particularly on readmission and LOS prediction. This may be due to the model’s focusing exclusively on sequential ICD codes—without incorporating aggregated hospital-level features such as ward transfers, demographics, or contact networks. These findings reinforce the importance of integrating both structured clinical codes and broader contextual features when modeling real-world hospital outcomes.}  \highlightc{One of the reasons for this may be the relatively coarse granularity of the ward-based contact measures and perhaps room or bed-based measures would have more impact, if that data was available.}

\subsubsection{CPE infection forecast}
\label{sec:results:cpe}

\highlight{In addition to forecasting patient outcome, we evaluated the feasibility of predicting CPE acquisition risk at admission by training models on screened patients, using only input features unrelated to HCAIs. Specifically, we excluded any downstream infection or outcome labels and constructed the prediction task using pre-admission features (e.g., patient history, current admission ward) and in-hospital contact network metrics derived from ward co-location. This setup simulates a real-world scenario in which hospitals must assess CPE carriage risk early using available admission and contact information, before any lab-confirmed diagnosis is made, before any clinical confirmation. We further leverage explainability methods to assess the influence of contact-based versus pre-admission features, identifying whether the model learns transmission-relevant signals from network structure.}

\begin{table*}[ht]
    \centering
    \caption{Benchmark performance of baseline and Transformer-based models on CPE risk estimation task using AUROC / AUPRC. All models evaluated on the same held-out test set using cross-validation setting.}
    \label{tab:benchmark_cpe_models}
    \resizebox{\textwidth}{!}{%
    \begin{tabular}{|l|cccc|}
        \hline
        \textbf{Model} & \textbf{AUROC} & \textbf{AUPRC}  & \textbf{Sensitivity}  & \textbf{Specificity}  \\
        \hline
        XGBoost                     & 0.755 (0.043) &  0.036 (0.019) & 0.075 (0.081) & 0.960 (0.070) \\
        CatBoost                     & 0.645 (0.078) & 0.027 (0.025) & 0.028 (0.020) & \underline{0.996} (0.002) \\
        LightGBM                     & 0.719 (0.092) & 0.030 (0.021) & 0.034 (0.028)  & 0.996 (0.005)  \\
        TabNet \cite{arik2021tabnet}   & 0.558 (0.083) & \textbf{0.111} (0.055)  & 0.125 (0.031) & 0.996 (0.004) \\
        ResNet \cite{gorishniy2021revisiting}  & \underline{0.773} (0.030) & 0.046 (0.030) & \underline{0.150} (0.125) & 0.969 (0.039) \\
        TabPFN \cite{hollmann_accurate_2025}  & 0.719 (0.051) & 0.022 (0.010) & 0.000 (0.000) & \textbf{1.000 (0.000)} \\
        TabTransformer \cite{huang2020tabtransformer}  & \textbf{0.781} (0.057) &  \underline{0.080} (0.017) & \textbf{0.425} (0.211) & 0.909 (0.071) \\
        CoreBEHRT \cite{odgaard2024core}       & 0.522 (0.127) & 0.015 (0.008) & 0.000 (0.000) & \textbf{1.000 (0.000)} \\
        \hline
    \end{tabular}
    }
\end{table*}

\highlight{As shown in Table~\ref{tab:benchmark_cpe_models}, Transformer-based models such as TabTransformer and ResNet outperform traditional boosting methods across AUROC and sensitivity, with TabTransformer achieving the highest AUROC (0.781) and  notably strong sensitivity (0.425), whereas most other models fall below 0.2, highlighting its robustness in identifying true positives in a highly imbalanced setting. }

\highlight{While AUPRC remains low due to extreme class imbalance, TabNet demonstrates a competitive edge in precision (AUPRC = 0.111), suggesting that attention-based tabular models may offer utility in identifying a smaller pool of high-risk individuals. Notably, CoreBEHRT underperforms due to its lack of access to the full feature set, reinforcing the need for aggregated representations beyond sequential clinical codes. These results demonstrate that early risk estimation of CPE is feasible using contact-informed features, paving the way for targeted screening and timely IPC intervention.}

\subsection{Explainable AI}
\label{sec:results:xai}

\subsubsection{Global Feature Importance}

\paragraph{Patient Outcome Tasks}

\sjh{To address our first research question—\textbf{\textit{“How do infection-related features impact on patient outcomes such as 30-day
readmission, in-hospital mortality, and future LOS?”}}, we analyze feature importance across all outcome tasks using IG. }

% % [trim={left bottom right top},clip]
% \begin{wrapfigure}{r}{0.5\textwidth}
%   \begin{center}
%     \vspace*{-0.5cm}\includegraphics[width=\linewidth,trim={2.8cm 0cm 3.2cm 1.1cm},clip]{Figures/new/importance/boxplot_cross_task_rankings_CM_averaged_Z225_B968_all_models_top5.pdf}
%   \end{center}
%     \caption{Impact of CPE-related ICD-10CM diagnosis codes on patient outcome modeling using IG. }
%   \label{fig:importance:cpe_icd}
%   \vspace*{-1cm}
% \end{wrapfigure}

\begin{wrapfigure}{r}{0.5\textwidth}
  \begin{center}
    \vspace*{-0.5cm}\includegraphics[width=\linewidth,trim={7.6cm 0cm 3.2cm 1.1cm},clip]{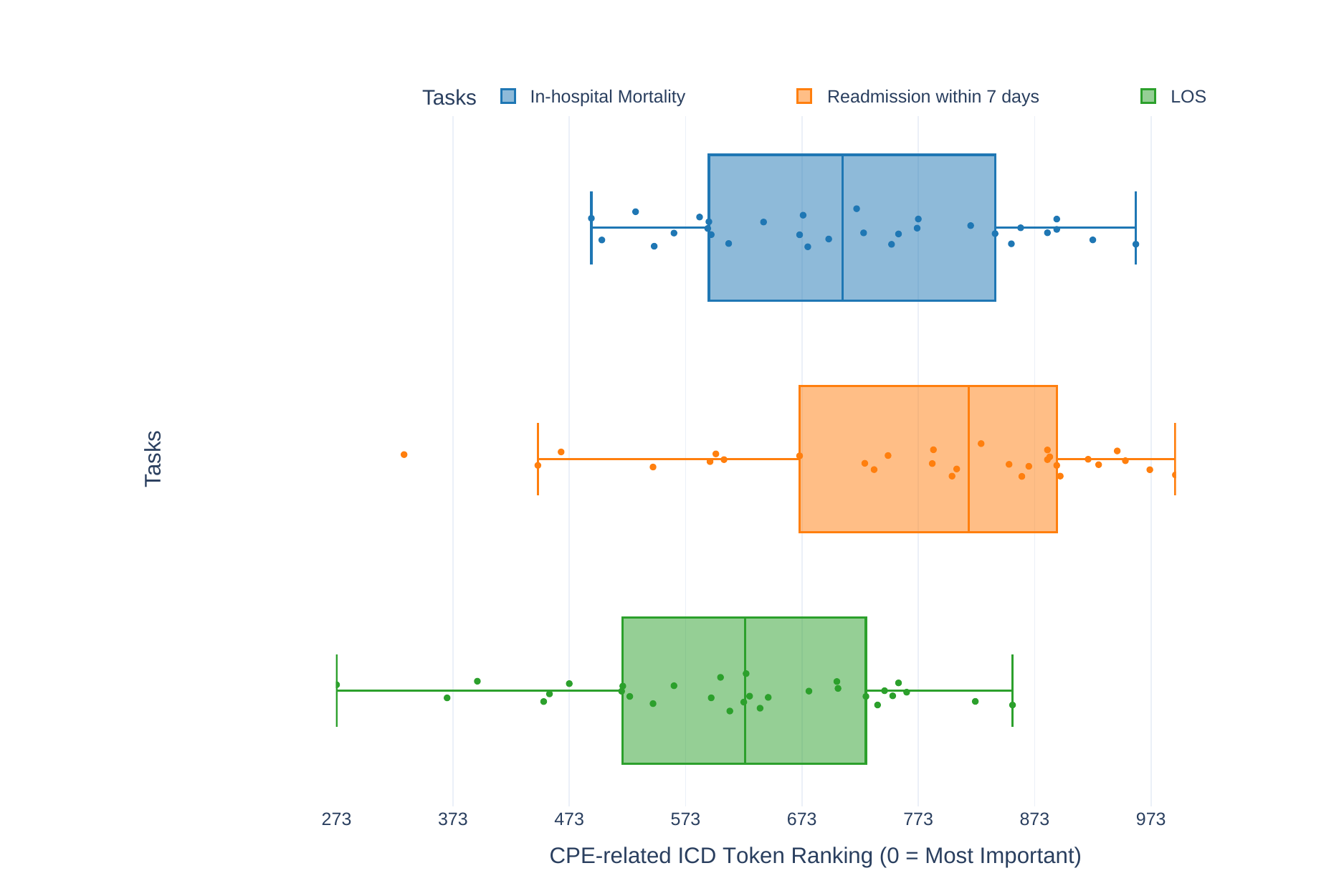}
  \end{center}
    \caption{Impact of CPE-related ICD-10CM diagnosis codes on patient outcome modeling using IG. }
  \label{fig:importance:cpe_icd}
  \vspace*{-1cm}
\end{wrapfigure}

\begin{figure}[ht]
    \centering
    \begin{subfigure}[t]{0.48\linewidth}
        \centering
        \includegraphics[width=\linewidth,trim={0.1cm 0.9cm 1.7cm 1cm},clip]{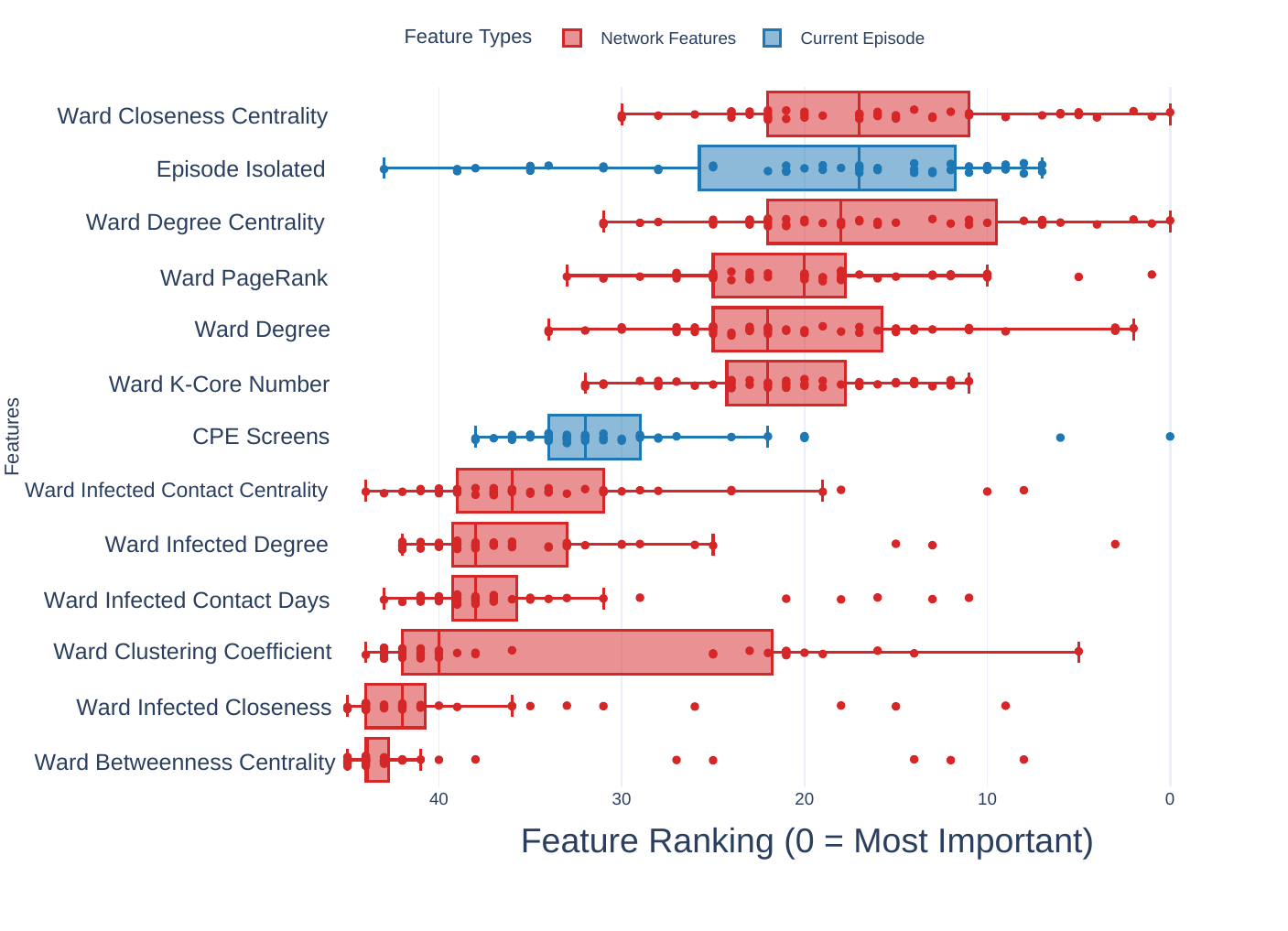}
        \caption{30-day Readmission}
    \end{subfigure}
    \begin{subfigure}[t]{0.48\linewidth}
        \centering
        \includegraphics[width=\linewidth,trim={0.1cm 0.9cm 1.7cm 1cm},clip]{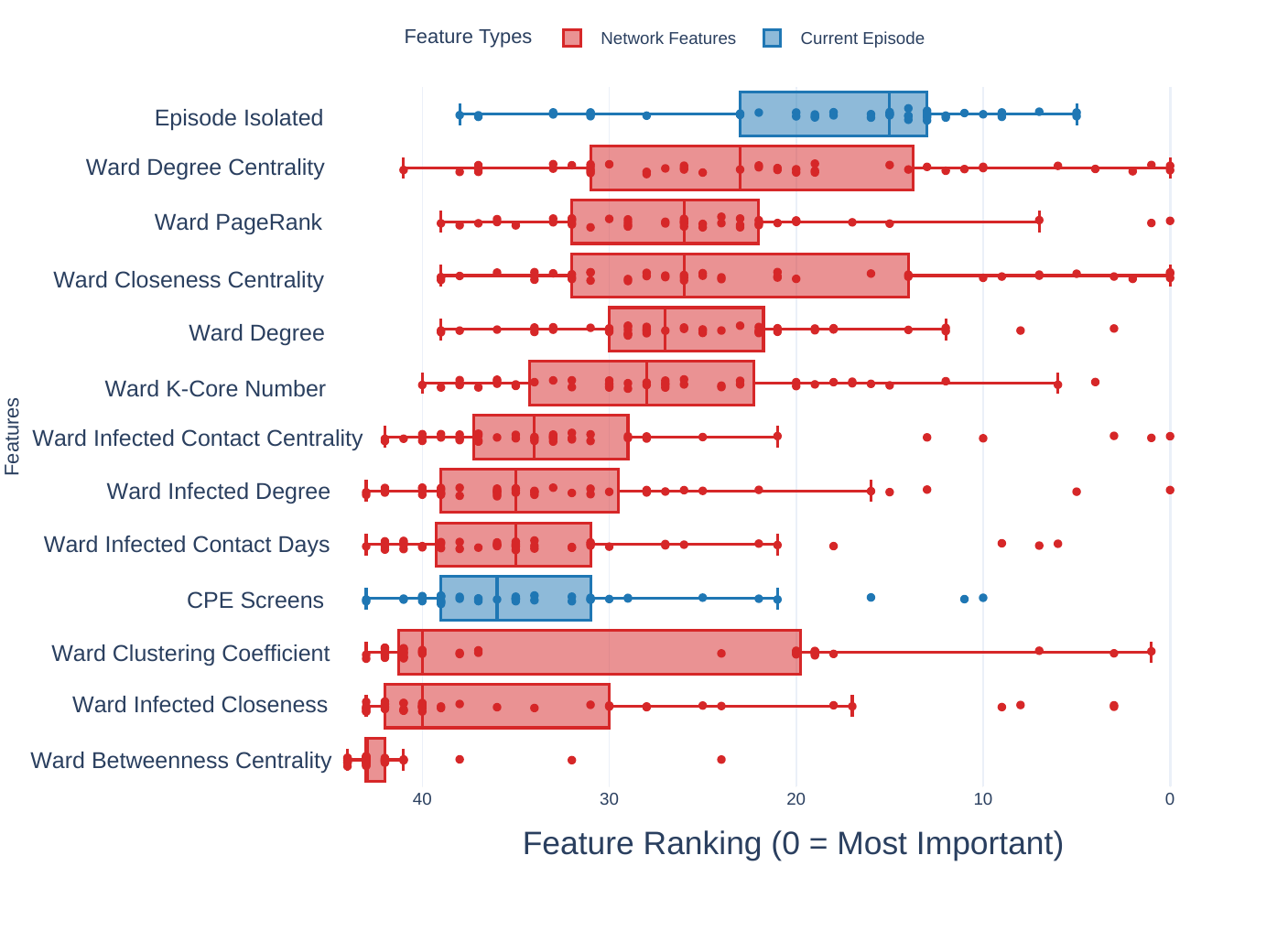}
        \caption{In-Hospital Mortality}
    \end{subfigure}
    \hfill
    \\
    
    \begin{subfigure}[t]{0.48\linewidth}
        \centering
        \includegraphics[width=\linewidth,trim={0.1cm 0.9cm 1.7cm 1cm},clip]{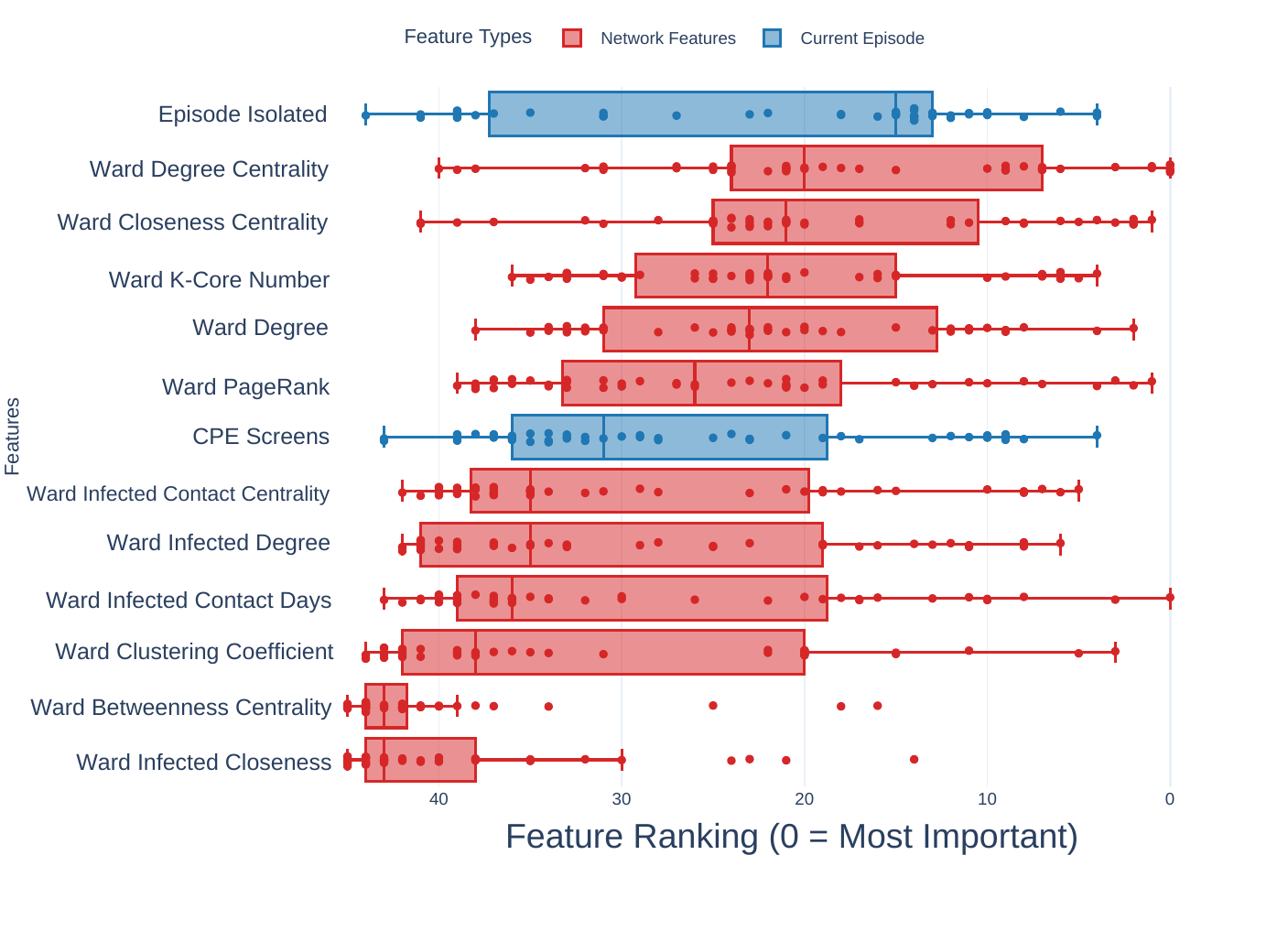}
        \caption{Length of Stay}
    \end{subfigure}
    \vspace{0.5em}
    \begin{subfigure}[t]{0.48\linewidth}
        \centering
        \includegraphics[width=\linewidth,trim={0.1cm 0.9cm 1.7cm 1cm},clip]{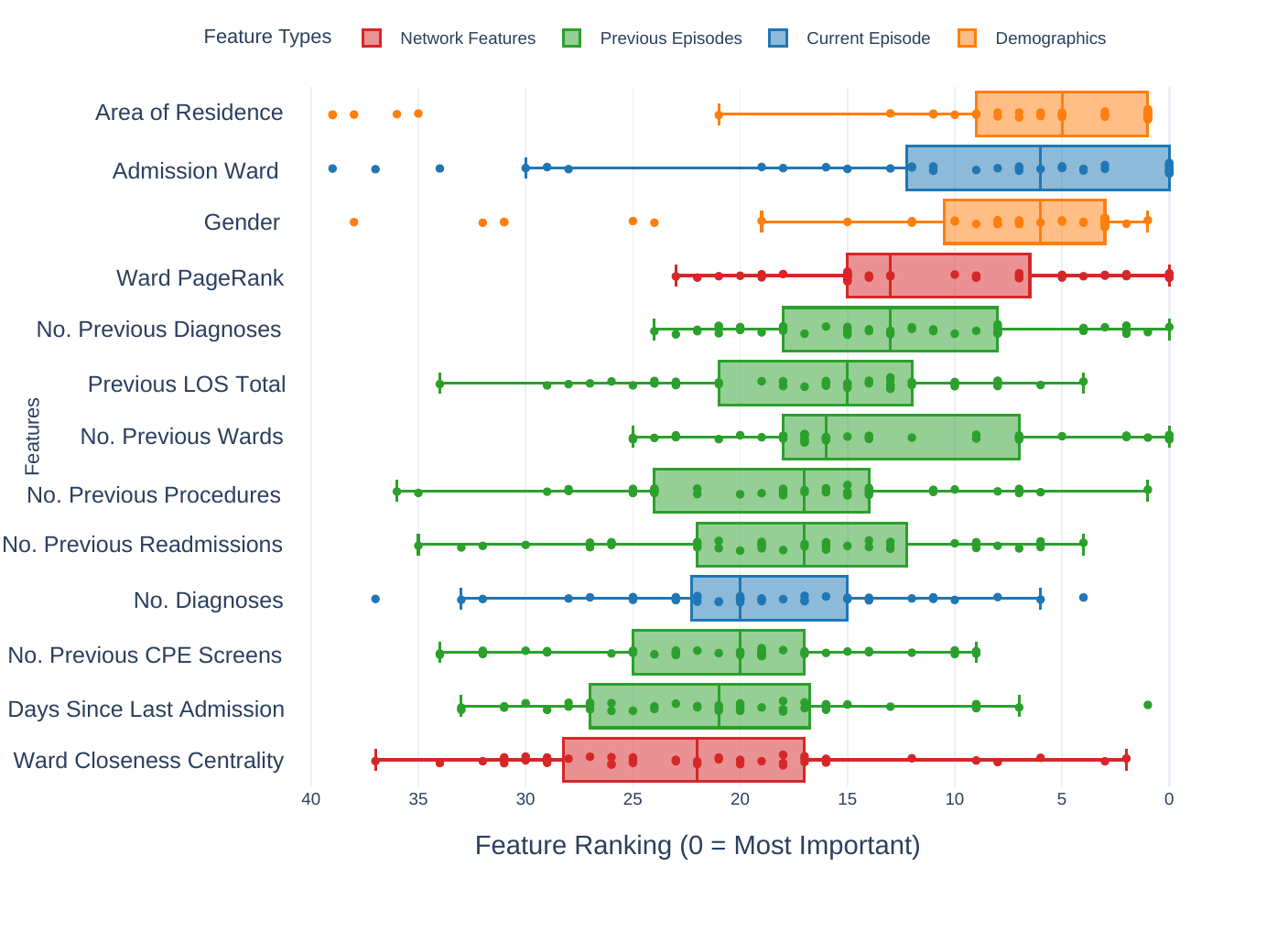}
        \caption{CPE Acquisition}
    \end{subfigure}
    
    \caption{Feature importance rankings based on Integrated Gradients for each prediction task. Shown in a), b) and c) are network-related and infection-related features (definitions can be found in Appendix). Whereas, d) demonstrates top predictive features of CPE acquisition from our benchmarked models. Lower ranks indicate higher importance. \crule[color1]{0.2cm}{0.2cm}: Demographics, \crule[color2]{0.2cm}{0.2cm}: Network features, \crule[color3]{0.2cm}{0.2cm}: Past Episode Features, \crule[color4]{0.2cm}{0.2cm}: Current Episode Features}
  \vspace*{-0.7cm}
    \label{fig:importance:global}
\end{figure}

\sjh{Figure~\ref{fig:importance:cpe_icd} and Figure~\ref{fig:importance:global} present the rankings of input features based on IG attribution scores, aggregated across five cross-validation folds and three model architectures (TabTransformer, ResNet, TabNet). Features are ranked by their median importance across models and folds, where lower ranks indicate greater predictive influence.}

\sjh{Firstly, we analyze the attribution scores of CPE-related clinical codes in Figure~\ref{fig:importance:cpe_icd} to assess their contribution to outcome prediction. On average, these ICD-10-CM codes rank relatively low—typically beyond position 650 out of approximately 1,000 total codes—indicating limited standalone impact on the model’s predictions. Among the outcomes, the length of stay (\texttt{"NEXT\_LOS"}) task shows slightly higher variability, with a few instances where CPE-related codes receive notably higher importance rankings. This suggests that while these features may influence predictions in specific contexts, their global contribution remains limited.}

% [trim={left bottom right top},clip]

\sjh{While patient contact network features were included to provide contextual information about intra-hospital dynamics, their overall contribution to outcome prediction appears moderate. For 30-day readmission, in-hospital mortality, and length of stay, variables such as \texttt{"Ward Closeness Centrality"}, \texttt{"Ward PageRank"}, \texttt{"Ward Degree Centrality"}, and \texttt{"Episode Isolated"} generally rank between 15 and 25 out of over 40 features, suggesting a mid-range importance across tasks.}

\sjh{These results indicate that contact network metrics offer supplementary value to the models, particularly in capturing structural aspects of patient flow. However, they are not among the most consistently influential features across the dataset. The wide interquartile ranges observed for some of these variables suggest that their relevance may vary by context or patient subgroup—potentially becoming more prominent in localized settings. Overall, their contribution appears to complement, rather than dominate, the predictive utility of other clinical and demographic features.}

\paragraph{CPE Infection Forecasting} 

\sjh{To address our second research question—\textbf{\textit{“Which patient features are the most predictive of CPE infection risk?”}}—we assess which features contributed most to CPE acquisition prediction using the same method and report in (Figure~\ref{fig:importance:global}d). The top-ranked features included \texttt{"Area of Residence"} and \texttt{"Admission Ward"}, suggesting that the model found contextual and ward-specific entry points informative, as found in the list of CPE risk factors in Table \ref{tab:cpe_risk_factors}. \texttt{"Ward PageRank"}, a network centrality measure, also ranked highly, indicating that ward-level contact structure may contain useful signals for identifying potential acquisition risk.}

\sjh{Additionally, several historical features—such as \texttt{"No. Previous Readmissions"}, \texttt{"Previous LOS Total"}, and \texttt{"No. Previous Diagnoses"}—consistently appeared among the top contributors. This reflects the model's use of longitudinal patient information when estimating risk. While these patterns highlight potentially important data signals, they represent model-driven associations and should not be interpreted as causal or clinical risk factors without further validation.}

\subsubsection{Embedding Projection}

\begin{wrapfigure}{r}{0.5\textwidth}
  \vspace*{-0.7cm}
  \begin{center}
    \includegraphics[width=\linewidth,trim={0.8cm 1.8cm 4.5cm 2cm},clip]{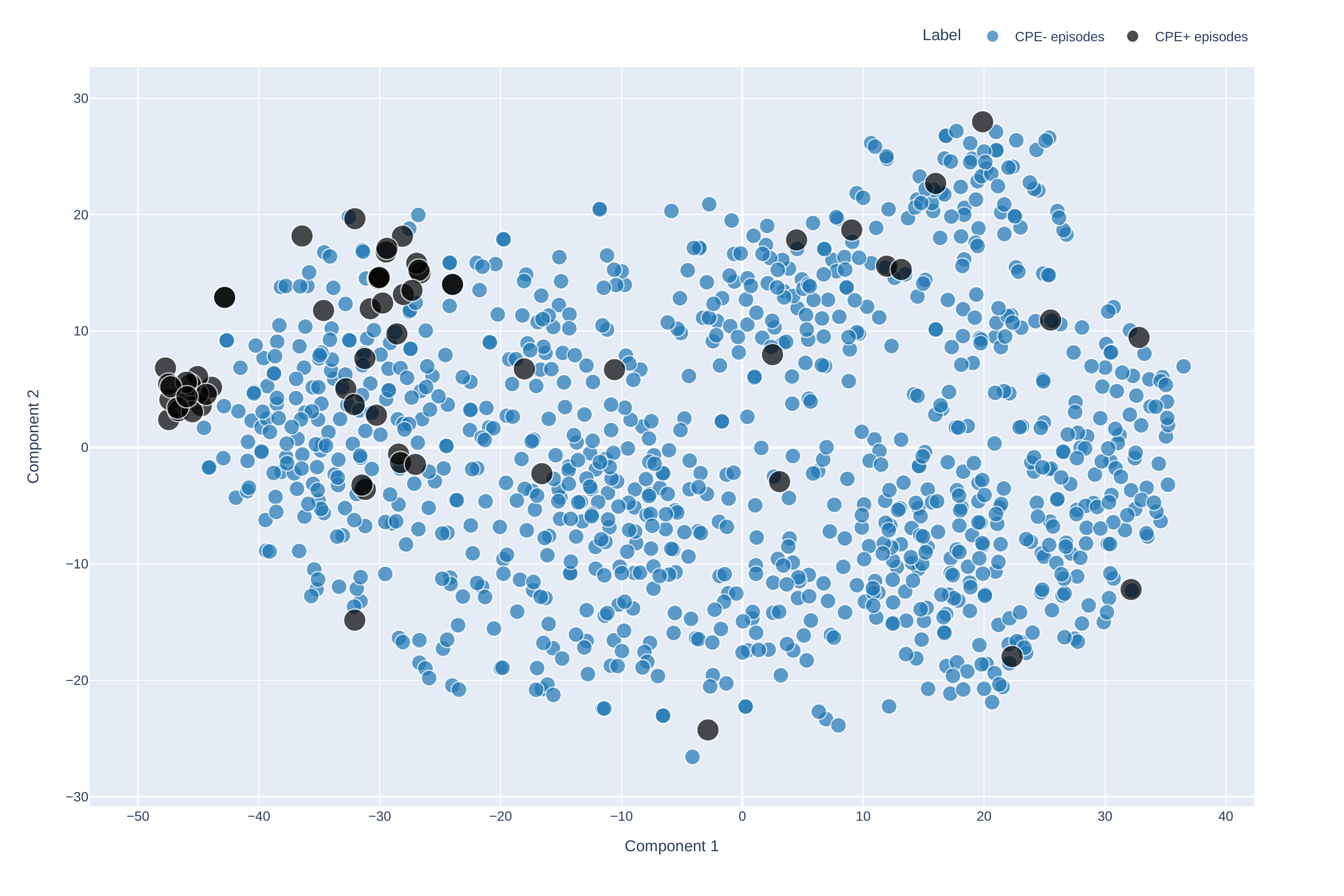}
  \end{center}
  \caption{t-SNE projection of learned patient episode embeddings from TabTransformer. Colors denote CPE status: \crule[color4]{0.2cm}{0.2cm} CPE-, \crule{0.2cm}{0.2cm} CPE+.}
  \label{fig:importance:episodeembedding}
  \vspace*{-0.7cm}
\end{wrapfigure}

\begin{figure}[ht]
    \centering
    \includegraphics[width=0.7\linewidth,trim={0.5cm 0 2cm 0},clip]{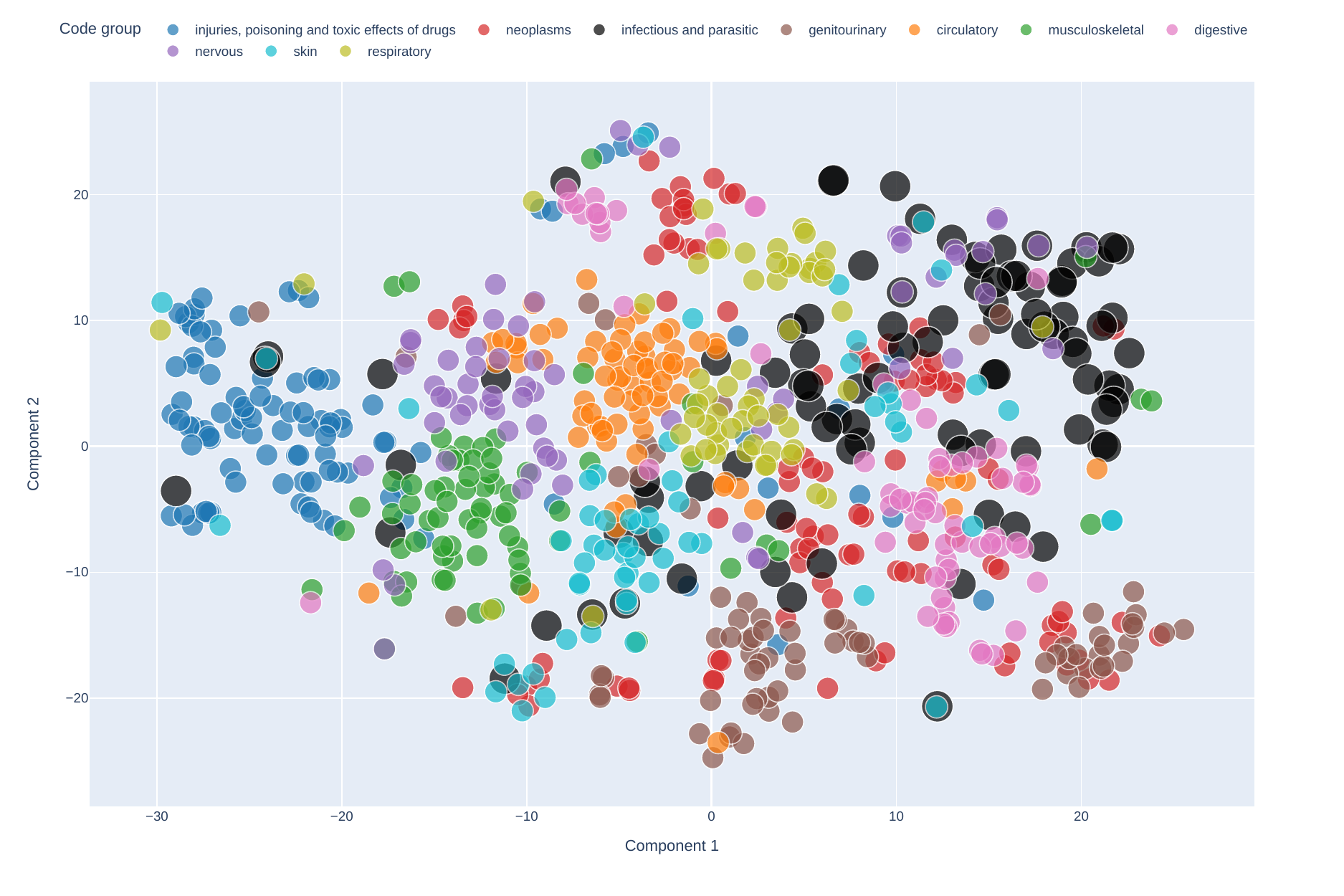}
    \caption{t-SNE projection of pretrained embeddings for the top 10 most frequent high-level ICD code groups, based on \cite{wu2022comparison}.}
    \label{fig:wordembedding}
  \vspace*{-0.7cm}
\end{figure}

Beyond population-level analysis, we explore cohort-level representations using embedding visualizations of both clinical codes and patient episodes, shown in Figure~\ref{fig:importance:episodeembedding} and Figures~\ref{fig:wordembedding}.

We project patient-level embeddings from the best-performing model in Table \ref{tab:benchmark_cpe_models} (TabTransformer) using t-SNE. As shown in Figure~\ref{fig:importance:episodeembedding}, episodes involving CPE-positive patients are visibly clustered together on one side of the embedding space. This occurs despite CPE status not being used as an explicit training label, indicating that the model implicitly learns meaningful clinical patterns related to CPE carriage. Such clustering highlights the model's ability to recognize between patient cohorts, proving its training capability.

Figure~\ref{fig:wordembedding} illustrates a t-SNE projection of pretrained embeddings for the ten most frequent ICD code groups in our dataset, following grouping conventions from \cite{wu2022comparison}. Codes are aggregated by their highest-level categories and projected into 2D space using t-SNE. The resulting plot reveals a clear separation between different diagnostic categories, suggesting that the embeddings capture semantic relationships among codes. The \texttt{"infectious and parasitic"} group, shown in black, forms a contiguous cluster, though some codes appear closer to other categories—indicating shared or overlapping characteristics. This visual structure supports the use of pretrained embeddings in representing clinical semantics and inter-code dependencies.

 Together, these visualizations suggest that deep learning models can capture latent representations that reflect real-world clinical structure and infection risks. Embedding projections offer a complementary perspective to traditional attribution methods by revealing global patterns in patient populations and code semantics. These findings support the model’s ability to detect infection-related characteristics even in the face of class imbalance, and demonstrate their value for visualizing patient flow and infection risk at scale—both essential for hospital IPC decision-making.

% [trim={left bottom right top},clip]
% \begin{figure}[ht]
%     \centering
%     \begin{subfigure}[t]{0.48\linewidth}
%         \centering
%         \includegraphics[width=\linewidth,trim={0.8cm 1.8cm 4.5cm 2cm},clip]{Figures/new/projections/tsne_visualization_tabtransformer_4.pdf}
%         \caption{TabTransformer}
%     \end{subfigure}
%     \hfill
%     \begin{subfigure}[t]{0.48\linewidth}
%         \centering
%         \includegraphics[width=\linewidth,trim={0.8cm 1.8cm 4.5cm 2cm},clip]{Figures/new/projections/tsne_visualization_tabnet_4.pdf}
%         \caption{TabNet}
%     \end{subfigure}
    
%     \vspace{0.5em}
    
%     \begin{subfigure}[t]{0.48\linewidth}
%         \centering
%         \includegraphics[width=\linewidth,trim={0.8cm 1.8cm 4.5cm 2cm},clip]{Figures/new/projections/tsne_visualization_resnet_4.pdf}
%         \caption{ResNet}
%     \end{subfigure}
%     \hfill
%     \begin{subfigure}[t]{0.48\linewidth}
%         \centering
%         \includegraphics[width=\linewidth,trim={0.8cm 1.8cm 4.5cm 2cm},clip]{Figures/new/projections/tsne_visualization_corebehrt_4.pdf}
%         \caption{CoreBEHRT}
%     \end{subfigure}

%     \caption{Embedding projection of patient visits learnt by different deep models. Colors denote the CPE status. \crule[color4]{0.2cm}{0.2cm}: CPE- episodes, \crule{0.2cm}{0.2cm}: CPE+ episodes}
%     \label{fig:importance:episodeembedding}
% \end{figure}

\subsubsection{Personalized explaination}

% \todo[inline]{
% 3. Do infection control interventions—such as isolation and screening—reduce the risk of poor outcomes or transmission?  In what way can xAI methods be employed for personalised investigation of individual case involving a CPE carrier?
% }

% \todo[inline]{
% Write about how much would save if apply this method as screening risk assessment \\
% Use personalized IG on unscreened patients and identify key features that might correlates with or missed by national standard criterias. \\
% --> if possible, visualize the attention models of sequences
% }

\begin{figure}[ht]
\centering
\includegraphics[width=1\textwidth,trim={0 0 2cm 2cm},clip]{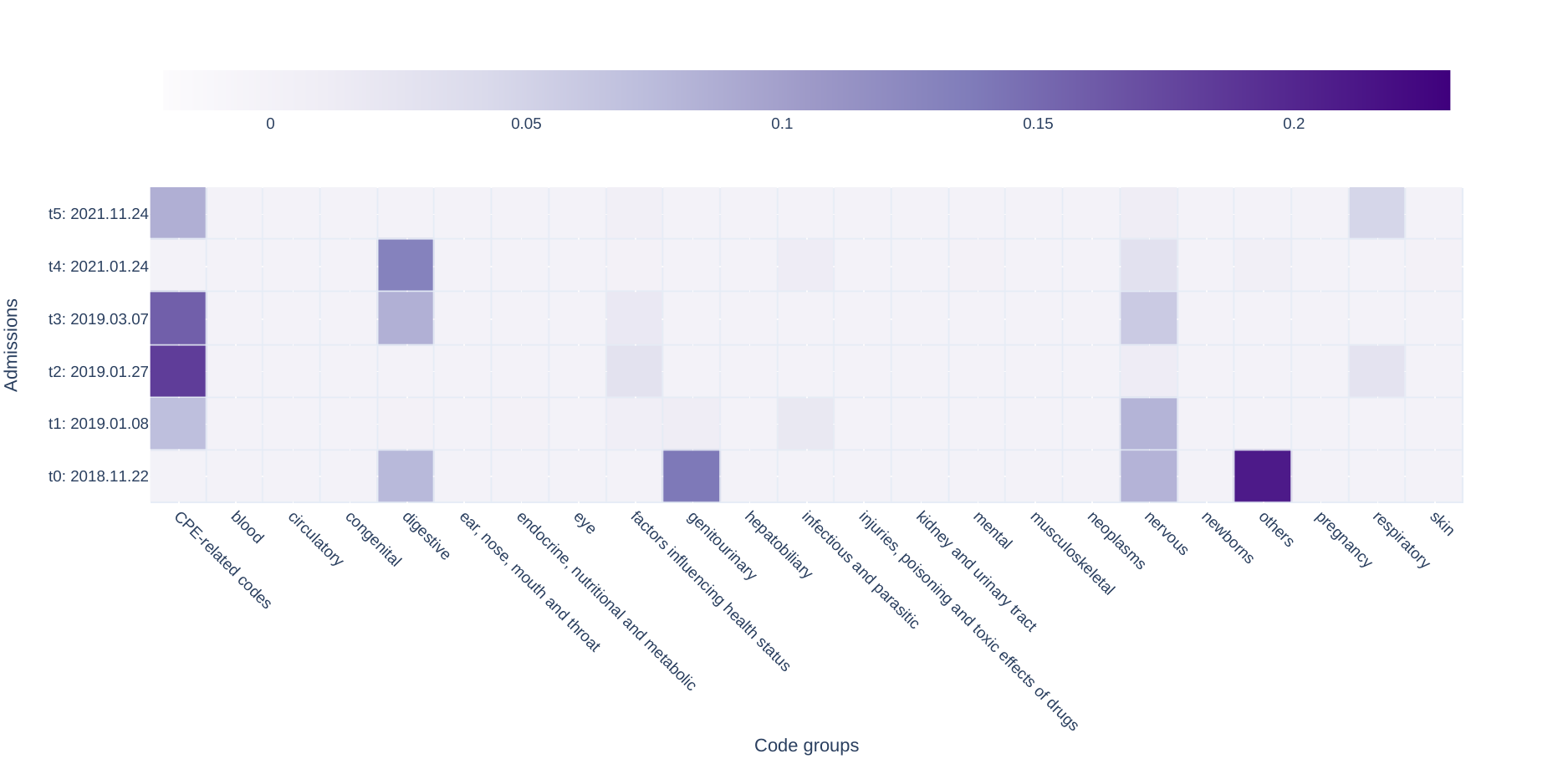}
\caption{Heatmap illustrating the importance of features for a randomly selected patient with multiple readmissions. }
\label{fig:integratedgradients}
  \vspace*{-0.7cm}
\end{figure}

In a recent study, Duell et al. proposed a modified version of IG tailored to capture temporal dynamics in patient records, enabling the analysis of evolving patient characteristics over time \cite{duell2023batch}. In contrast, our approach treats each hospitalization episode independently, as our features are aggregated as static data rather than modeled as temporal sequences. To illustrate the potential impact of CPE on frequent readmissions, we present a case study of a representative patient in Figure~\ref{fig:integratedgradients}. While this individual example suggests that CPE-related risks may contribute to multiple short-term readmissions, it should be emphasized that this observation cannot be generalized due to the limited number of CPE-positive cases in the dataset.

Figure~\ref{fig:integratedgradients} visualizes the attention dynamics across sequential episodes for the selected patient, highlighting how the model assigns importance values to different ICD code groups over time. The heatmap’s color intensity indicates the contribution of each group to the predicted risk of readmission for the following episode.

% [trim={left bottom right top},clip]

In this sample, we identify three distinct hospitalization phases in the patient’s trajectory: episodes \textit{t0–t3}, \textit{t4}, and \textit{t5}. In the first phase, increased IG values to CPE-related codes coincides with a cluster of closely spaced readmissions, suggesting possible CPE acquisition and microbiological diagnosis. Additionally, in most phases, diagnoses related to the nervous system and digestive appear more prominently. This example demonstrates the potential of attribution methods to support personalized risk profiling and retrospective investigation of recurrent readmission drivers.

\section{Discussion}
\label{sec:discussion} 

\highlight{This study presents a practical application of deep tabular learning and Transformer-based models for assessing CPE-associated risks using real-world EMR data. Rather than proposing novel architectures, our contribution lies in developing a flexible explainable AI (XAI) framework that integrates diverse patient features to support infection prevention and control (IPC) efforts.}

\highlight{Transformer-based models, particularly TabTransformer, consistently outperformed traditional baselines in AUROC across all predictive tasks. However, performance on precision-recall metrics remained modest due to the dataset's inherent class imbalance. \highlightc{Since this imbalance is inherent to many rare event risks like CPE, this may limit practical applications without new ways to aggregate data such as multi-microbial event studies such as TREE \cite{rock2025electronic}.} For the CPE acquisition task, TabTransformer also achieved the highest sensitivity score, suggesting its potential utility in identifying high-risk patients who may otherwise be missed under rule-based screening criteria. While not a diagnostic tool, such a model could assist IPC staff in prioritizing patients for early screening, \highlightc{potentially based on admission information (\texttt{"Area of Residence"}, \texttt{"Admission Ward"}), and historical indicators such as number of prior admissions and diagnoses}, thereby reducing manual workload and enabling more proactive surveillance.}

\highlight{To address data imbalance, we applied targeted mitigation strategies, including focal loss and balanced sampling, alongside the introduction of contact network-derived features. These variables provide spatial and contextual insight by capturing patient movement and potential exposure pathways within the hospital. Although these network features did not consistently rank among the top predictors across all tasks, their presence in mid-tier importance and variability across folds suggest that they may offer complementary value in specific scenarios.}

\sjh{Our flexible use of Integrated Gradients for interpretability allows deeper inspection of model predictions. Feature attribution analyses revealed that CPE-related clinical codes had limited contribution to outcome prediction—likely a reflection of the small number of confirmed cases in the dataset. Nonetheless, several literature-supported risk factors for CPE acquisition (e.g., prior admissions, length of stay, admission ward) were ranked highly in the acquisition task, demonstrating that the model can recover meaningful associations from daily EMR data. Additionally, embedding visualizations suggest that the model learns distinguishable representations for CPE-positive patients, further supporting its potential for cohort-level stratification.}

\sjh{Finally, in practice, this framework can be integrated as a risk-scoring module into hospital IPC systems. Given EMR and patient location data, the system outputs: (1) impact of CPE prevalence on patient outcomes (2) risk of CPE acquisition and (3) correlated features behind each prediction. Moreover, an interpretability dashboard can flag patients who may benefit from targeted screening or preemptive isolation—especially those falling outside national criteria. \highlightc{For example, a patient with frequent short admissions across multiple wards and high contact network exposure might be flagged for screening, even if they do not meet criteria such as previous inpatient care in the last twelve months or ICU admission}. This bridges the gap between prediction and action, aligning model capabilities with IPC goals.}

% %%%%%%%%%%%%%%%%%%%%%%%%%%%%%%%%%%%%%%%%%%

% %%%%%%%%%%%%%%%%%%%%%%%%%%%%%%%%%%%%%%%%%%

\section{Limitations \& Future Works}
\label{sec:limitations}

The short timeframe of our dataset further exacerbates data imbalance, with CPE-positive cases representing only 0.2\% of all recorded episodes. Data imbalance is a major concern when using AI technologies to deal with real-world AMR datasets since it can cause ML models to output inconsistent and unreliable results \cite{ali2023artificial}. The insufficient sample size hinders the DL model's ability to effectively learn and draw robust conclusions about HAI-related risks. Fitzpatrick et. al \cite{fitzpatrick2020using} also points out the requirement for a high-quality representative dataset to develop accurate models for HCAI surveillance. In our dataset, the quality of the clinical coding is highly dependent on the information in the discharge summary that were collected, which we found to be inconsistent among different coders, but effects can be mitigated by generalizing the codes. Moving forward, it is crucial to address this challenge by prioritizing the collection of more extensive and high-quality datasets. This will enable the development of more accurate and reliable models, ultimately improving clinical risk management and patient outcomes. Exploring aggregation across related infection events (e.g., multi-microbial studies) could also help improve signal for model training. Nonetheless, we believe that our proposed model and XAI techniques hold promise for application in other desired patient groups.

\highlight{A promising direction to address these limitations lies in the use of autoregressive longitudinal modeling. Our recent work, NOSOS \cite{pham2025nosos}, demonstrates that foundation models designed for token-by-token generation can learn temporal dynamics in patient health trajectories more robustly, even in imbalanced settings. These models treat EMR data as a generative sequence and have shown resilience to missing data and rare outcomes. Integrating such generative approaches with infection prevention and control tasks may enable more accurate unsupervised forecasting of HCAIs over time, particularly in resource-constrained or low-prevalence settings.}

\highlight{Another avenue worth exploring is the refinement of TabPFN-based models \cite{hollmann_accurate_2025}. Although TabPFN performed surprisingly well under the small-data regime, it is currently limited by constraints on input dimensionality and sample size. With recent developments in scaling few-shot learning for tabular data, future work could investigate whether adapting or retraining TabPFNs on larger, healthcare-specific priors can further improve their generalization and performance in imbalanced infection datasets.}

Another important consideration is data privacy and generalizability. Our analysis is constrained to a single Irish hospital, and patient privacy protections prevented linking patient data across hospital systems. This limitation introduces a risk of mislabeling—for instance, if patients were readmitted elsewhere without corresponding records. Therefore, future work should explore multi-institutional datasets, such as MIMIC-III/IV \cite{johnson2016mimic, johnson2023mimic}, to validate our findings in broader settings and ensure more robust generalization across clinical contexts. Moreover, while AUROC values were strong across tasks, the relatively modest precision-recall performance highlights a limitation for practical applications, where high precision is critical for identifying small groups of high-risk patients. Precision-oriented methods such as calibrated thresholds, cost-sensitive learning, or ensemble strategies should be explored to better support real-world clinical decision-making.

\section{Conclusion}
\label{sec:conclusion}

This study presents an explainable AI framework for investigating CPE-related risks and outcomes using structured EMR data from an Irish acute hospital. We benchmarked a suite of traditional and Transformer-based models, finding TabTransformer to consistently outperform baselines across multiple clinical prediction tasks, particularly in AUROC and sensitivity for CPE acquisition. The integration of diverse patient features—including ward history, contact networks, and clinical codes—allowed for a comprehensive analysis of infection-related outcomes. Throughout, explainability analysis was conducted via the flexible use of Integrated Gradients and embedding visualizations, enabling transparent model interpretation. Our study addressed four key research questions:

\vspace{0.5em}
\noindent
\textbf{(RQ1)} Our findings show that infection-related features, including historical hospital exposure and admission context, are moderately associated with patient outcomes such as readmission, mortality, and LOS. Contact network variables—designed to capture in-hospital exposure—ranked mid-tier in feature importance, suggesting contextual relevance in certain scenarios but not dominating model predictions. CPE diagnosis codes, due to data scarcity, had limited impact on outcome forecasting.

\vspace{0.5em}
\noindent
\textbf{(RQ2)} For predicting CPE acquisition risk, the most influential features included \texttt{"Area of Residence"}, \texttt{"Admission Ward"}, and historical indicators such as number of prior admissions and diagnoses. These features align with known CPE risk factors in the literature, indicating that the model successfully recovers associations from routine EMR data. Network centrality measures like \texttt{"Ward PageRank"} also ranked highly, reflecting the potential value of structural exposure information.

\vspace{0.5em}
\noindent
\textbf{(RQ3)} Our results show that explainable AI techniques, particularly Integrated Gradients and embedding projections, enable multi-level insights into the model’s handling of CPE-positive patients. At the cohort level, CPE-positive episodes form distinguishable clusters in the embedding space, suggesting the model can capture latent characteristics associated with this subgroup. At the individual level, attribution visualizations reveal how diagnostic histories and contact-related features contribute to predictions, offering interpretable signals that may support more tailored IPC interventions. While these patterns are not clinically validated, they highlight the potential of XAI to support case-level audit and decision-making in real-world hospital workflows.

Overall, our framework offers a practical, interpretable pipeline for infection risk estimation that could be deployed in hospital IPC systems to guide early screening, risk scoring, and resource allocation.
\backmatter

% \bmhead{Supplementary information}

% If your article has accompanying supplementary file/s please state so here. 

% Authors reporting data from electrophoretic gels and blots should supply the full unprocessed scans for key as part of their Supplementary information. This may be requested by the editorial team/s if it is missing.

% Please refer to Journal-level guidance for any specific requirements.

% \bmhead{Acknowledgements} Not applicable

\section*{Declarations}

\subsection{Funding}
This research was conducted with the financial support of Taighde Éireann-Research Ireland under Grant Agreement No. 13/RC/2106\_P2 at ADAPT, the Research Ireland Centre for AI-Driven Digital Content Technology at DCU funded through the Research Ireland Research Centres Programme. For the purpose of Open Access, the author has applied a CC BY public copyright licence to any author-accepted manuscript version arising from this submission.

\subsection{Competing interests}
 The authors declare that they have no competing interests. 

\subsection{Ethics approval and consent to participate}

The study was conducted using a real-world anonymized dataset that was shared as part of the ARK risk management research project. Our study was approved by the St James Hospital's institutional review board (TUH/SJH REC Project ID: 0291) and by Dublin City University Research Ethics Committee, Dublin, Ireland (DCUREC/2021/118). The requirement for informed consent was waived by the Ethics Committee of St James Hospital’s institutional review board because of the retrospective nature of the study and was also considered unnecessary according to national regulations at the time of the research. All methods were carried out in accordance with relevant guidelines and regulations, and in line with the ethical principles of the WMA Declaration of Helsinki (as amended in 2024), with strict safeguards to ensure patient confidentiality and privacy.

\subsection{Abbreviations}
AI: Artificial intelligence; DL: Machine Learning; DL: Deep Learning; CPE: Carbapenemase-Producing Enterobacteriacea; HCAI:  Healthcare-Associated Infection; EMRs: Electronic Medical Records; XAI: Explainable AI; AUROC: Average area under the receiver operating characteristic curve; AMR: antimicrobial resistance;  IG: Integrated Gradients; HIPE: Hospital Inpatient Enquiry;  ICD-10CM: International Classification of Diseases Codes version 10 Clinical Modification; IPC: Infection Prevention and Control.

\subsection{Consent for publication}
Not applicable.

\subsection{Data and materials availability}

The datasets generated or analysed during the current study are not publicly available due to restrictions imposed by institutional ethics approvals and data protection regulations but are available from the corresponding author on reasonable request. The code and reproducibility guidelines supporting the analyses are publicly available at \url{https://github.com/kaylode/carbapen}.

\subsection{Author contributions}
MKP: Involved in all aspects of this study. MKP, MC and MB: Conceptualization. MEW, UG, DB, BO, CB, DC: Data acquisition and interpretation
MKP, TTM, MC, RB, MEW, UG, DB, BO, CB, DC, NM, MB: revision of the manuscript. MKP, MEW, UG, DB, BO, CB, DC: Result validation and interpretation. All authors read and approved the final manuscript.

\subsection{Acknowledgements}
Not applicable.

\bibliography{sn-bibliography}% common bib file


\end{document}